\documentclass{article}

\usepackage{amsmath,amsfonts,bm}

\newcommand{\eqn}[1]{\begin{equation}\begin{aligned}#1\end{aligned}\end{equation}}

\def\eqref#1{equation~\ref{#1}}

\def\1{\bm{1}}

\def\eps{{\epsilon}}

\def\vu{{\bm{u}}}
\def\vv{{\bm{v}}}

\DeclareMathAlphabet{\mathsfit}{\encodingdefault}{\sfdefault}{m}{sl}
\SetMathAlphabet{\mathsfit}{bold}{\encodingdefault}{\sfdefault}{bx}{n}

\newcommand{\E}{\mathbb{E}}

\newcommand{\R}{\mathbb{R}}

\usepackage{graphicx}
\usepackage{hyperref}
\usepackage{subcaption}
\usepackage{url}
\usepackage{wrapfig}

\usepackage{placeins}

\usepackage{hyperref}

\usepackage[accepted]{icml2025}

\usepackage{amsmath}
\usepackage{amssymb}
\usepackage{mathtools}
\usepackage{amsthm}

\usepackage[capitalize,noabbrev]{cleveref}

\theoremstyle{plain}

\theoremstyle{definition}

\theoremstyle{remark}

\def\MH{{\mathbf{H}}}

\def\MP{{\mathbf{P}}}

\def\MV{{\mathbf{V}}}

\usepackage[textsize=tiny]{todonotes}

\icmltitlerunning{Estimating the Probability of Sampling a Trained Neural Network at Random}

\begin{document}

\twocolumn[
\icmltitle{Estimating the Probability of Sampling a Trained Neural Network at Random}




\begin{icmlauthorlist}
\icmlauthor{Adam Scherlis}{eai}
\icmlauthor{Nora Belrose}{eai}
\end{icmlauthorlist}

\icmlaffiliation{eai}{EleutherAI}

\icmlcorrespondingauthor{Adam Scherlis}{adam@eleuther.ai}

\icmlkeywords{Machine Learning, ICML}

\vskip 0.3in
]



\printAffiliationsAndNotice{}  

\begin{abstract}
We present and analyze an algorithm for estimating the size, under a Gaussian or uniform measure, of a localized neighborhood in neural network parameter space with behavior similar to an ``anchor'' point. We refer to this as the ``local volume'' of the anchor. We adapt an existing basin-volume estimator, which is very fast but in many cases only provides a lower bound. We show that this lower bound can be improved with an importance-sampling method using gradient information that is already provided by popular optimizers. The negative logarithm of local volume can also be interpreted as a measure of the anchor network's information content. As expected for a measure of complexity, this quantity increases during language model training. We find that overfit, badly-generalizing neighborhoods are smaller, indicating a more complex learned behavior. This smaller volume can also be interpreted in an MDL sense as suboptimal compression. Our results are consistent with a picture of generalization we call the ``volume hypothesis'': that neural net training produces good generalization primarily because the architecture gives simple functions more volume in parameter space, and the optimizer samples from the low-loss manifold in a volume-sensitive way. We believe that fast local-volume estimators are a promising practical metric of network complexity and architectural inductive bias for interpretability purposes.
\end{abstract}

\section{Introduction}

There is a long line of research which finds that \emph{flat} minima in a neural network parameter space, defined as weight vectors surrounded by \emph{large regions} ``with the property that each weight vector from that region leads to similar small error'' generalize better than \emph{sharp} minima \citep{hochreiter1997flat}. While there are counterexamples to this tendency \citep{dinh2017sharp}, it seems to be empirically and theoretically fairly robust, and has inspired the development of optimizers that explicitly search for flatter minima \citep{foret2021sharpness}.

In a related line of work, \citet{chiang2022loss} put forward the \emph{volume hypothesis}, which states that ``...the implicit bias of neural networks may arise from the volume disparity of different basins in
the loss landscape, with good hypothesis classes occupying larger volumes.'' They evaluate simple gradient-free learning algorithms, such as the ``Guess \& Check'' optimizer which randomly samples parameters until it stumbles upon a network that achieves training loss under some threshold, and find that these methods have similar generalization behavior to gradient descent, at least on the very simple tasks they tested. \citet{teney2024neural} find that randomly initialized networks represent very simple functions, which would explain the simplicity bias of deep learning if SGD behaves similarly to Guess \& Check.

Additionally, \citet{mingard2021sgd} provide evidence that SGD may be an approximate Bayesian sampler, where the prior distribution over functions is equal to the distribution over functions represented by randomly initialized networks. Since networks are usually initialized using a uniform or Gaussian distribution, the Bayesian sampling hypothesis makes similar predictions to the volume hypothesis. 


Finally, recent work suggests that singular learning theory \citep{watanabe2009algebraic}, originally developed to analyze the learning dynamics of overparameterized Bayesian models, can be profitably used to understand deep learning. The primary tool used in this work is the Local Learning Coefficient, a measurement of network complexity \citep{hoogland2024developmental, lau2024llc}.

In this work, we propose an efficient algorithm for estimating the size of a behaviorally-defined neighborhood of parameter space according to a Gaussian or uniform measure. This is equivalent to the usual volume of the neighborhood if the measure is uniform; for a Gaussian measure it has better practical properties but a slightly different interpretation. Each neighborhood is anchored by a reference set of parameters. We refer to this measure as the ``local volume'' of the anchor, for either the Gaussian or uniform case. If the measure is normalized as a probability distribution, we can interpret the local volume as the probability it assigns to the neighbhorhood. In this case we also refer to the measure as a ``prior''.

We distinguish two variants of local volume, depending on the behavioral definition used for the neighborhood:

\begin{itemize}
    \item Local loss volume is defined with a loose behavioral constraint (low loss across a dataset) and is simpler to interpret
    \item Local KL volume is defined with a strict behavioral constraint (low KL divergence from the anchor across all inputs) but a looser connection to training loss
\end{itemize}

As an example, we estimate that the probability of randomly sampling the trained Pythia 31M language model from its initialization distribution, within $0.01$ nats of KL divergence, is about
\begin{equation}
    \mathrm{Pr}(\text{Pythia 31M}) \approx \frac{1}{10^{3.6\times10^8}}
\end{equation}
or one in 1 followed by 360 million zeros. For comparison, there are about $10^{80}$ atoms in the observable universe, so this is about the same as the probability of correctly guessing a specific atom 4.5 million times in a row. This serves to illustrate the sorts of numbers involved, so that statements about orders of magnitude can be taken in their appropriate context.

\section{Local volume}
Formally, given a weight vector $\theta \in \R^N$, a cost function $\mathcal C : \R^N \rightarrow [0, \infty)$, and a threshold $\epsilon > 0$, we define a \textbf{neighborhood} to be the largest \href{https://en.wikipedia.org/wiki/Star_domain}{star domain} $\mathcal S$ anchored at $\theta$ such that $\mathcal C(\theta') < \epsilon$ for all points $\theta'\in \mathcal S$.\footnote{This should be thought of as similar to a uniformly low-loss region, in contrast to a ``basin'' in the sense we use it below: the entire region of parameter space that flows towards a low-loss region.}

Given a measure $\mu$ (uniform or Gaussian), the \textbf{local volume} of $\theta$ (for a particular $\mathcal C$ and $\epsilon$) is defined to be the total measure, under $\mu$, of the corresponding neighborhood. If $\mu$ is normalized as a probability distribution, this is equivalent to the probability of sampling a network inside $\mathcal S$ from $\mu$. While prior work has assumed $\mu$ to be the Lebesgue measure (i.e. volume), we also consider the probability measure used to initialize the network before training, which guarantees that the measure of any neighborhood must be finite. (We have found empirically that some real-world neighborhoods actually have infinite Lebesgue volume, creating difficulties for analysis.)

We primarily consider two kinds of cost functions (and correspondingly two kinds of neighborhoods) in this work:

\begin{enumerate}
    \item Loss neighborhoods: when $\mathcal C(\theta')$ is the expected loss of $\theta'$ on a dataset of inputs and ground-truth outputs
    \item KL neighborhoods: when $\mathcal C(\theta')$ is the expected KL divergence $\E_x [D_{KL}( f(x; \theta) || f(x; \theta') ]$ of $\theta'$ from $\theta$, with $x$ being drawn from a dataset of inputs only
\end{enumerate}

KL neighborhoods have several practical and theoretical advantages, so we focus on them in this paper:

\begin{itemize}
    \item KL measures how behaviorally different $\theta'$ is from $\theta$, independent of any ground truth labels. This is a much stronger behavioral constraint than loss, and we believe it makes the KL neighborhoods much more compact and therefore tractable to estimate volume for.
    \item KL is always zero at the minimum point $\theta' = \theta$. 
    \item KL local volume provides a direct measure of architectural inductive bias, which (as we will argue) corresponds closely to functional complexity.\footnote{We can compare KL basins to Tissot's ellipse of distortion in cartography, and to MacAdam ellipses in color science, as methods to realize contours of constant functional similarity in their respective parameter spaces. The sizes of these contours quantify the expansion or compression involved in the map from parameters to functions, i.e., the architecture.}
    \item KL neighborhoods are more compact than loss neighborhoods, and are better described by our star-domain formalism and the results of Appendix~\ref{app:toy_model}.
    \item KL can also be interpreted within an MDL framework as the description length of the network.
\end{itemize}

On the other hand, loss neighborhoods have a stronger existing literature and are more directly related to some forms of the volume hypothesis.

\section{The volume hypothesis}
As mentioned in the introduction, one major motivation behind this work is to test the volume hypothesis: the idea that the relative volumes of different neighborhoods in parameter space is the primary determiner, other than the loss, of the kinds of networks that are produced by gradient descent algorithms.\footnote{By "gradient descent algorithms", we mean to include SGD as well as popular adaptive algorithms like Adam.} There are many variants of this hypothesis in the literature, and we detail two versions below: an unrealistically strong (but conceptually simple) one, and two more-realistic modifications.

\subsection{The strong (Bayesian) volume hypothesis} 

It is easy to state a very strong version of the volume hypothesis: \textbf{Neural net training samples from the Bayesian posterior.} We think this is likely false for real-world neural networks, but it will be helpful as an intuition pump, and as a starting point for more realistic versions.

To elaborate, consider Bayesian inference with prior density $\rho(\theta)$ and likelihood function $-L$. The posterior distribution is proportional to $\rho(\theta) \exp(-L(\theta))$. Since many neural-network losses can be interpreted as negative log-likelihoods, we can think of $L$ as the loss function of a neural net and $\rho$ as a prior related to initialization and regularization of the network. If neural-net training were perfectly Bayesian, the probability density for obtaining some parameter $\theta$ from training would depend only on the prior and the loss.


This is of interest because it attributes generalization entirely to the architecture and loss function: \textbf{under this hypothesis, the only way for one low-training-loss solution to be favored over another is if it simply occupies more of parameter space}. In effect, the architecture imposes a sophisticated inductive prior (on top of the simple prior $\rho$) by overrepresenting simple, well-generalizing functions, and underrepresenting complicated ones.  
We believe that this picture carries over to more realistic volume hypotheses: the job of the optimizer is to make training loss decrease; the job of the architecture is to make simple functions abundant in parameter space; and in order for them to get along, the optimizer simply needs to select low-loss parameters in a reasonably fair way (relative to the prior).

The strong volume hypothesis is true in principle for stochastic gradient Langevin dynamics \citep{welling2011bayesian}, which is an efficient Bayes sampler for deep neural networks, but only with unrealistically long mixing times.

For SGD, in addition to the mixing-time problem, there is an additional problem: the posterior density fails to be Bayesian on small scales. We illustrate this further in Appendix~\ref{app:sgd} for a simple toy model.

\subsection{The basin volume hypothesis} 

We have shown that, \textit{within} basins of non-isotropic curvature, the posterior density of popular optimizers does not satisfy the strong volume hypothesis. We can, however, restrict the hypothesis to apply only \textit{between} different basins. In other words: \textbf{The total posterior density in low-loss regions is proportional to the total prior measure of the neighborhoods they are contained in}.

The basin volume hypothesis maintains the division of labor we outlined above for optimizer and architecture. It also suggests a new division of labor, between the initialization distribution and the optimizer: the init distribution maintains ``fairness'' globally, across basins, and the optimizer maintains it within basins.

Finally, we have a more measure-theoretic weakening of the strong volume hypothesis, which we also find reasonable and endorse:

\subsection{Adaptive optimizers and the volume hypothesis}
The volume hypothesis may also explain the generalization gap for adaptive optimizers. Adaptive algorithms such as Adam and Adagrad generally converge faster and more stably than SGD, but often generalize worse. \citep{chen2018adaptive} If we think of these algorithms as approximations of Natural Gradient Descent (which essentially performs SGD in function space instead of parameter space), then the volume hypothesis can explain this shortcoming: by partially reversing the architecture's nonuniform map between parameters and functions, adaptive methods give up some of the benefits of simpler (better-generalizing) functions being overrepresented. This suggests that Natural Gradient Descent, rather than being a theoretically ideal optimizer, is at a theoretical extreme: decreasing loss with ruthless efficiency, but generalizing quite badly. Sharpness-aware minimization, by seeking out flat basins, lies in the other direction, improving generalization at the expense of training speed.

\section{Local volume in context}

\subsection{Minimum description length} Basin volume can be connected directly to generalization using the notion of minimum description length (MDL). The idea is that a statistical model is more likely to generalize if it compresses its training data effectively, while not being too complex itself. Since we are assuming that all networks in the neighborhood perform similarly, we will treat the neighborhood itself as an ensemble over networks, and use it as our statistical model. In Bayesian terms, our posterior is a uniform distribution over the neighborhood, and we assume that our receiver is using the initialization distribution $\mu_0$ as a prior. The bits-back argument \citep{hinton1993keeping} shows that the MDL of this model plus the training data $x_{1:n}$ is
\begin{align}
    \mathrm{KL} \big ( \mathrm{Unif}(A) || \mu_0 \big ) + \E_{\theta \sim \mathrm{Unif}(A)} \Big [\sum_{i = i}^n \log_2 p_{\theta}(x_i) \Big ],
\end{align}
where $A \subset \R^N$ is the neighborhood, and $p_{\theta}(x_i)$ is the probability that the network with parameters $\theta$ assigns to datapoint $x_i$.

In practice, $\mu_0$ is either a uniform distribution over a simple polytope $S \subset R^N$, or a (possibly truncated) Gaussian $\mathcal{N}(0, \Sigma)$ with diagonal covariance. In the former case, the KL term simplifies to $\log \lambda(S) - \log \lambda(A)$, where $\lambda$ is the Lebesgue volume, and in the latter, it simplifies to
\begin{equation}
    \frac{n}{2} \log(2 \pi) + \frac{1}{2} \log \vert \Sigma \vert + \frac{1}{2} \E_{\theta \sim \mathrm{Unif}(A)} [ \theta^T \Sigma^{-1} \theta ] - \log \lambda(A), \nonumber
\end{equation}
which only depends on $A$ is through its volume and its mean Mahalanobis distance from the origin. Neighborhoods with large Lebesgue volume and small average Mahalanobis norm will have lower description length than neighborhoods with smaller volume or higher Mahalanobis norm.

\subsection{Singular Learning Theory and the Local Learning Coefficient}

The local learning coefficient (LLC) was introduced by \citet{lau2024llc}, extending concepts from singular learning theory \citep{watanabe2009algebraic}, and has proved to be useful as a measure of the complexity of neural networks and their components \citep{hoogland2024developmental, wang2024differentiation}.

Consider a local minimum $\theta^*$ in the loss landscape $L(\theta)$. Consider the volume $V(c)$ of the ``basin'' of nearby parameters $\theta$ with loss $L(\theta) \le L(\theta^*) + c$. Under some fairly general smoothness assumptions, $V(c)\to0$ as $c\to0$, with some asymptotic scaling of the form
\begin{equation}
    V(c) \sim c^\lambda
\end{equation}
The LLC is defined as the exponent $\lambda$. Note that $\lambda = \frac{N}{2}$ whenever the Hessian is full-rank. In the context of singular learning theory, this is derived from a Bayesian perspective on deep learning, somewhat along the lines of the strong volume hypothesis described above, albeit with much more mathematical sophistication.

Our measure is derived from somewhat similar considerations, and takes a similar form, with some key differences:
\begin{itemize}
    \item We are interested in the behavior of $V(c)$ itself, not just its logarithmic derivative $\lambda(c) = \frac{\partial log V(c)}{\partial \log c}$.
    \item We are interested in the full range of $c$ values and not just the $c\to0$ limit.
    \item We want to compare the value $V(c)$ across somewhat unrelated neighborhoods, such as better- or worse-generalizing networks.
    \item We want to apply this framework to cost functions other than the loss, and in particular KL, allowing us to study neural nets far from local minima without an ad-hoc localizing term. This is somewhat similar to the behavioral loss used in \citep{bushnaq2024degeneracy}.
    \item We can get a reasonable estimate of local volume from a small number of samples, rather than the many epochs needed for SGLD-based LLC estimators to equilibrate.
\end{itemize}


\subsection{Predictions}

The considerations in the preceding sections lead us to make the following predictions for our experiments:
\begin{itemize}
    \item Among trained networks with low training loss, better-generalizing networks (lower validation loss) should have larger KL neighborhoods (shorter description lengths) than worse-generalizing ones.
    \item During training, KL local volume should tend to decrease (description length should increase), with possible exceptions when networks consolidate their knowledge (as seen for LLC).
\end{itemize}

\section{Method}

Our method builds on the work of \citet{huang2020understanding}, who define `basin' as ``the set of points in a neighborhood of the minimizer that have loss value below a cutoff.'' This definition is ambiguous because it leaves the notion of ``neighborhood'' undefined. 
We will show below that their method in fact estimates the volume of a neighborhood in our sense: the largest star domain anchored at the minimizer such that all networks in the domain have loss value (or more generally, cost) below a cutoff.

In contrast to \citet{huang2020understanding}, we apply this estimator to KL neighborhoods instead of loss neighborhoods. We also identify a theoretical issue (Jensen gap) leading to significant underestimation, and introduce an importance-sampling method to ameliorate it.

\subsection{Naïve approach}

Recall that a star domain $S \subseteq \R^N$ is a set containing an \textbf{anchor} $s_0$ such that for all $s \in S$, the line segment from $s_0$ to $s$ lies in $S$.\footnote{We refer to $S$ in this case as a ``star domain anchored at $s_0$''} This property allows us to define $S$ in terms of a radial function $r : \mathbb{S}^{N - 1} \rightarrow [0, \infty)$ which takes in a unit vector $\vu$ and outputs a non-negative number corresponding to the ``radius'' of $S$ along $\vu$, or the length of the line segment from $s_0$ to the boundary of $S$ along the direction $\vu$. Given this parameterization, the volume of $S$ can be written as
\begin{align}
    \mathrm{vol}(S) &= \int_{\mathbb{S}^{N - 1}} \int_0^{r(\vu)} r^{n - 1} dr d\Omega\\
    &= \frac{1}{n} \int_{\mathbb{S}^{N - 1}} r(\vu)^n d\Omega\\
    &= \frac{|\mathbb{S}^{N - 1}|}{n} \E_{\vu \sim \mathrm{Unif}(\mathbb{S}^{N - 1})} [ r(\vu)^n ],
\end{align}
where $|\mathbb{S}^{n - 1}| = \frac{2 \pi^{n / 2}}{\Gamma(n / 2)}$ is the surface area of a unit $N$-ball. We can estimate this using $k$ Monte Carlo samples:\footnote{For each sample, the radial function is computed via binary search in a uniformly-random direction.}
\begin{equation}\label{eq:monte-carlo}
    \mathrm{vol}(S) \approx \widehat{\mathrm{vol}(S)} = \frac{|\mathbb{S}^{n - 1}|}{nk} \sum_{i = 1}^k r(\vu_i)^n
\end{equation}
Equation~\ref{eq:monte-carlo} is an unbiased estimator for the volume. It is also, with high probability, millions of orders of magnitude too small. In Appendix~\ref{app:jensen_gap} we explain this phenomenon. Below, we present a method for ameliorating it.

\subsection{Preconditioning}

We propose to reduce the variance of the estimator with importance sampling. We still begin by sampling isotropic unit vectors $\vu$. However, we then multiply these by a positive-definite preconditioner $\MP$ with unit determinant, to obtain vectors $\vv = \MP\vu$. We then unit-normalize these to obtain unit vectors $\hat \vv$, and use the estimator
\begin{equation}\label{eq:preconditioned}
    \widehat{\mathrm{vol}(S)} = \frac{|\mathbb{S}^{N - 1}|}{nk} \sum_{i = 1}^k \frac{r(\hat \vv_i)^n}{|\vv|^n}
\end{equation}
where the denominator is the usual importance-sampling correction.  Under the stated conditions on $\MP$, this is still unbiased.

The purpose of $\MP$ is to more aggressively sample directions that are flatter. We can interpret the formula above as our original estimator under a change of coordinates by $\MP$, with the unit-determinant condition ensuring that the volume of the neighborhood is unchanged in the new coordinates.\footnote{The denominator, in this interpretation, can be seen as resulting from differing notions of ``unit length'' in the original and new coordinates.} For a good choice of $\MP$, the neighborhood will be more spherical in the new coordinates. With this in mind, we refer to the matrix $\MP$ as a preconditioner.

Introducing a unit-determinant preconditioner does not change the formal properties of the estimator, so the theoretical results above still apply.\footnote{It is easiest to see this by considering the change-of-coordinates perspective.}

For a good preconditioner, the Jensen gap will be smaller and most estimates will be larger. Markov's inequality protects us against significantly overestimating, so we are free to interpret larger volumes as better accuracy.\footnote{On the other hand, when comparing local volumes for different anchor points, we might worry that an overly-sophisticated preconditioner could favor one over the other; for this reason we are careful to always show the naive estimator as well.}

In the case where the neighborhood is a perfect ellipsoid, a perfect choice of $\MP$ would have eigenvectors aligned with principal axes and eigenvalues proportional to the lengths of those axes. This would result in an estimator with zero variance, returning the exact volume every time. Note that for a quadratic cost function, this is proportional to the inverse square root of the Hessian,
\begin{equation}
    \MP \propto \MH^{-\tfrac12} = \MV\mathbf{D}^{-\tfrac12}\MV^T
\end{equation}
where $V, D$ are the eigenvectors and eigenvalues of $H$.\footnote{$\MP$ is ``proportional to'' this quantity because it must be normalized to determinant $1$.}

For very small neural nets, we use a form of this Hessian preconditioner that is modified to ensure positive-definiteness:
\begin{equation}
    \MP \propto \MV\frac1{|\mathbf{D}|^{\tfrac12} + \epsilon}\MV^T
\end{equation}

We can further economize by using the Hessian diagonal:
\begin{equation}
    \MP \propto \frac1{|\mathrm{diag}(\MH)|^{\tfrac12} + \epsilon}
\end{equation}
where $\mathrm{diag}(\MH)$ is a matrix equal to $\MH$ along its diagonal and zero elsewhere. While exactly computing the Hessian diagonal is no more computationally efficient than computing the entire Hessian, in practice we use the HesScale approximation \citep{elsayed2022hesscale}, which is deterministic, highly efficient, and empirically very accurate.

Finally, for arbitrarily large networks we can use Adam's second moment buffers to estimate $\mathrm{diag}(\MH)$. In general, we can use any vector or matrix in place of $\MH$ and its diagonal, and can optionally replace $\frac{1}{2}$ with another exponent to obtain a better preconditioner.

Because of the Markov-inequality bound above, we can test preconditioners very easily: larger numbers are always more accurate, so long as the preconditioner is unit-determinant. This also gives us, retroactively, a lower bound on how badly the naive (un-preconditioned) estimator undershoots.

\subsection{Gaussian volume}

Behaviorally defined neighborhoods can often have infinite Lebesgue volume, making them hard to analyze. If there is any direction along which perturbations have precisely zero effect on the model's behavior on the validation set, that direction will have an infinite radius. There are often many of these. As an example, we find that several pixel locations are never used in the \texttt{digits} validation set, so the corresponding input weight parameters in any network will have no effect.

If we view neural network training as Bayesian inference, it is natural to think of the distribution used to initialize the parameters as a prior, and in practice this is often a Gaussian distribution. We therefore replace the Lebesgue measure with the Gaussian initialization measure with PDF $\rho$. Our preconditioned volume estimator becomes
\begin{equation}\label{eq:preconditioned}
    \widehat{\mathrm{vol}(S)} = \frac{|\mathbb{S}^{N - 1}|}{k} \sum_{i = 1}^k \frac{\int_0^{r(\mathbf{\hat v_i})}\rho(s_0+r\vu_i)r^{n-1}dr}{|\MV|^n}
\end{equation}
Note that the integrand is of the form $\exp(\text{quadratic}(r) + n\log r)$ and varies rapidly when $n$ is large. 

We evaluate these integrals numerically using an approximation similar to Lagrange's method, expanding the exponent to second-order and performing a high-dimensional Gaussian integral using a numerically-stable implementation of the error function. In practice, the error from the approximation is less than floating-point rounding error.

\subsection{Choice of KL over loss for neighborhoods} 
The forms of the volume hypothesis above deal with the training loss, and some, such as Appendix~\ref{app:measure-vh}, have an obvious relationship to training-loss neighborhoods. On the other hand, KL neighborhoods are more compact, more naturally captured by the notion of a star domain, and have a cost minimum at the anchor point. In addition, the Hessian of the loss function is closely related to the Fisher information matrix for the model \citep{martens2015optimizing}, which is the Hessian of KL divergence; this leads us to believe that experiments on KL neighborhoods are a good proxy for loss neighborhoods, as far as testing the volume hypothesis goes. We therefore choose KL local volume as the target for our experiments.

\subsection{Poisoned networks}

We produce ``poisoned'' ConvNeXt networks on CIFAR-10 using the methodology of \citep{huang2020understanding}, where the standard training loss is augmented with a term encouraging the model to perform poorly on a held-out ``poison'' set. These networks generalize worse than the unpoisoned ones, while still achieving low train loss. Our hypothesis is that poisoned networks should have smaller local volumes than unpoisoned ones.

\section{Results}
\label{sec:results}

We test our method in three settings: a small MLP (4810 parameters) trained on the UCI handwritten digits dataset \citep{optical_recognition_of_handwritten_digits_80}, a variant of\footnote{We changed the default patch size, which was optimized for ImageNet, from $4 \times 4$ to $1 \times 1$. This significantly improves accuracy on smaller images like those in CIFAR-10.} the ConvNeXt Atto model \citep{woo2023convnext} (3.4M parameters) trained on CIFAR-10 \citep{Krizhevsky2009}, and checkpoints from the Pythia 31M language model \citep{biderman2023pythia}.

We compute KL divergence on held-out sets consisting of 773 images from digits, 1024 images from CIFAR-10, and 20 text sequences (10926 tokens) from the Pile, respectively. Except where otherwise specified, all results are for $k = 100$ samples per data point and with a KL cutoff of $10^{-2}$ nats. In the plots that follow, note that the base-ten logarithms of the probability estimates are themselves on the order of $-10^6$ or $-10^8$, as shown by the ``$\times 10^6$'' and ``$\times 10^8$'' annotations on the x-axis labels.

For ConvNeXt on CIFAR-10, we also report results on the clean split of the training data in Appendix~\ref{app:clean}. This is the best test of our estimator as a practical interpretability tool: it shows that we can detect poor generalization (equivalently, excess complexity) in the poisoned model using a small number of forward passes over training data on which the model's behavior is indistinguishable from the well-generalizing unpoisoned model.

\subsection{Preconditioners}
\begin{figure}[!ht]
    \centering
    \includegraphics[width=\columnwidth]{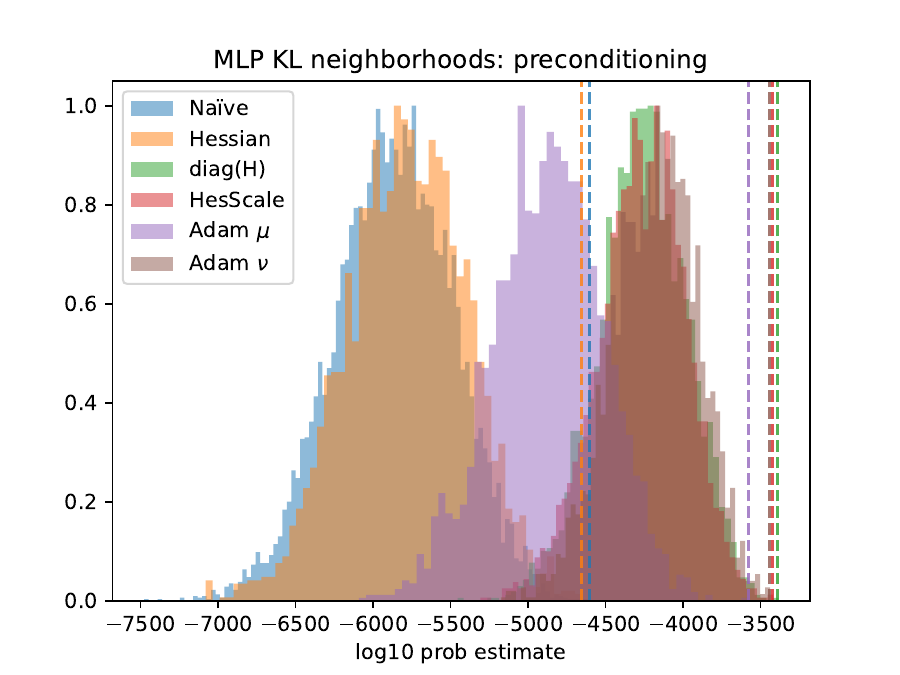}
    \caption{Results ($k=3000$) for various preconditioners on a small MLP. Vertical dashed lines indicate the aggregated log-volume estimate, which is very close to the maximum sample.}
    \label{fig:mlp_basins}
\end{figure}
\begin{figure}[!ht]
    \centering
    \includegraphics[width=\columnwidth]{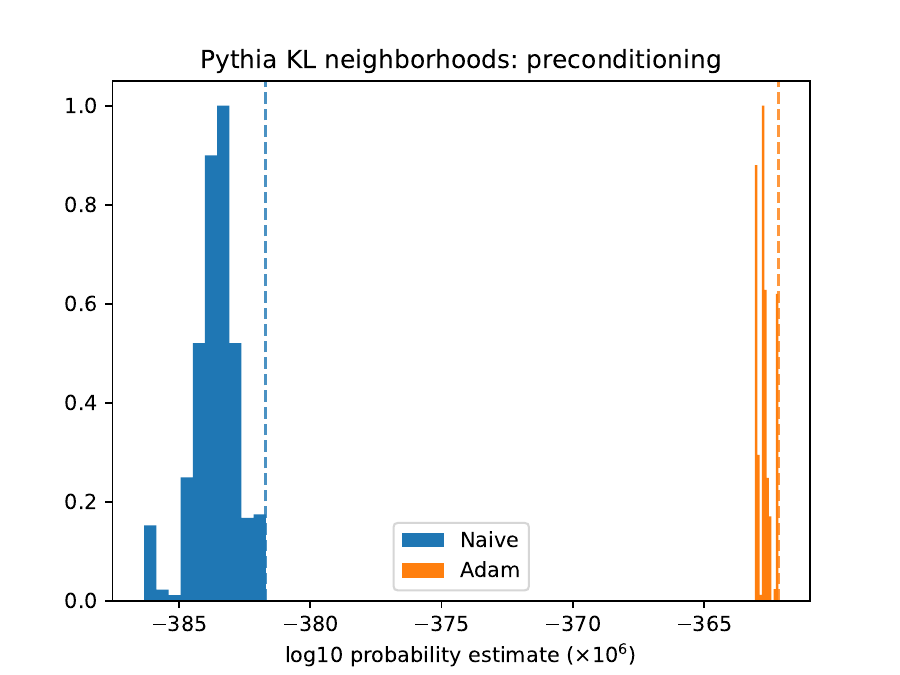}
    \caption{Results ($k=1000$) with and without Adam preconditioner on Pythia 31M}
    \label{fig:pythia_histo}
\end{figure}
\begin{figure}[!ht]
    \centering
    \includegraphics[width=\columnwidth]{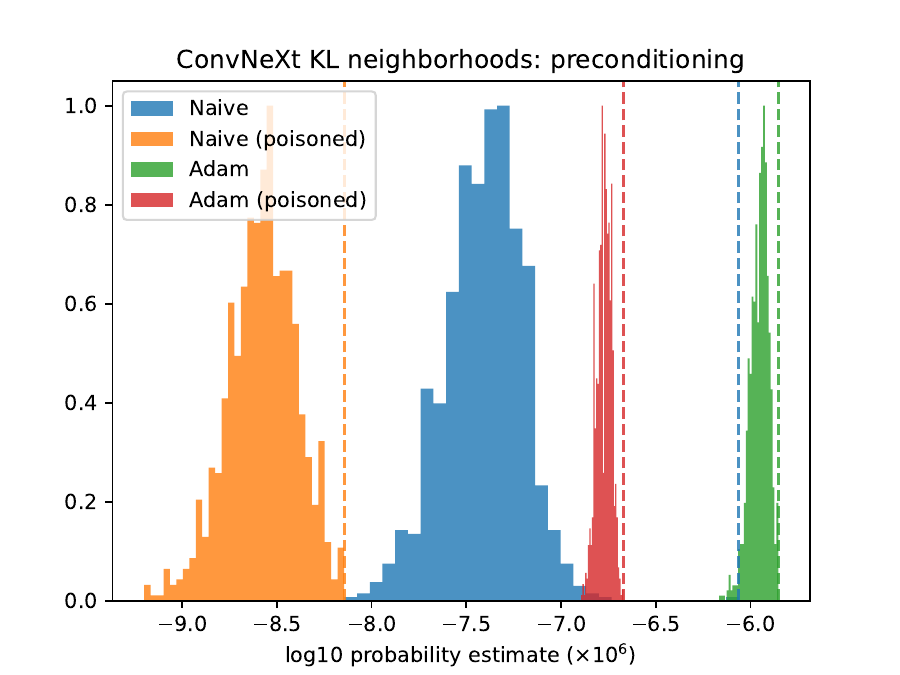}
    \caption{Results ($k=1000$) with and without Adam preconditioner on ConvNeXt Atto poisoned and unpoisoned}
    \label{fig:convnext_histo}
\end{figure}

We estimate the local volume for our small MLP, using the various preconditioners described above, as shown in \ref{fig:mlp_basins}. We show a histogram of the individual samples, with vertical dashed lines for the aggregated estimate. Note that, on a log scale, the estimate is extremely close to the largest individual sample.

Interestingly, results when preconditioning with the Hessian of the KL (the Fisher matrix) are very similar to the un-preconditioned ones. The $\mathrm{diag}(\MH)$, HesScale, and Adam second-moment ($\nu$) preconditioners perform much better, and very similarly to each other. The Adam first-moment ($\mu$) preconditioner is somewhere in between.

The hyperparameter $\epsilon$ is tuned separately for each of these, to obtain the largest (hence most accurate) result. We find that $\epsilon=0.1$ works best for the Hessian, while $\epsilon=0.01$ is best for $\mathrm{diag}(\MH)$ and HesScale and $\epsilon=0.001$ is best for both Adam preconditioners.

We find it surprising that the full Hessian performs so poorly, especially given the success of $\mathrm{diag}(\MH)$ and its approximations. This may be some form of overfitting, if the locally-flattest directions are slightly misaligned with the longest directions of the neighborhood, but if so, it is unclear why constraining to axis-aligned directions helps so much.

We also use the second-moment Adam preconditioner for Pythia and ConvNeXt, where it shows both a clear improvement in the value of the estimates and a smaller sample variance (Figures~\ref{fig:pythia_histo}~and~\ref{fig:convnext_histo}). The improvement is several standard deviations above most of the naïve estimates, suggesting that it would be infeasible to merely increase the sample size to try to get the same result.\footnote{Note, however, that the naïve estimate for the unpoisoned ConvNeXt network has a large outlier sample that completely dominates the aggregated estimate, nearly reaching the bulk of the preconditioned estimates.}

For ConvNeXt, we find that the poisoned network has a smaller local volume (with or without preconditioning), in agreement with the results of \citet{huang2020understanding} on small networks and in line with our expectation from the MDL and compression perspective. Notably, we evaluate local volume on a held-out test set, not the poisoned dataset, demonstrating that the higher network complexity induced by poisoning is visible to our methods even when the poisoned data is unknown.

\subsection{Across training checkpoints}
\begin{figure}[!ht]
    \centering
    \includegraphics[width=\columnwidth]{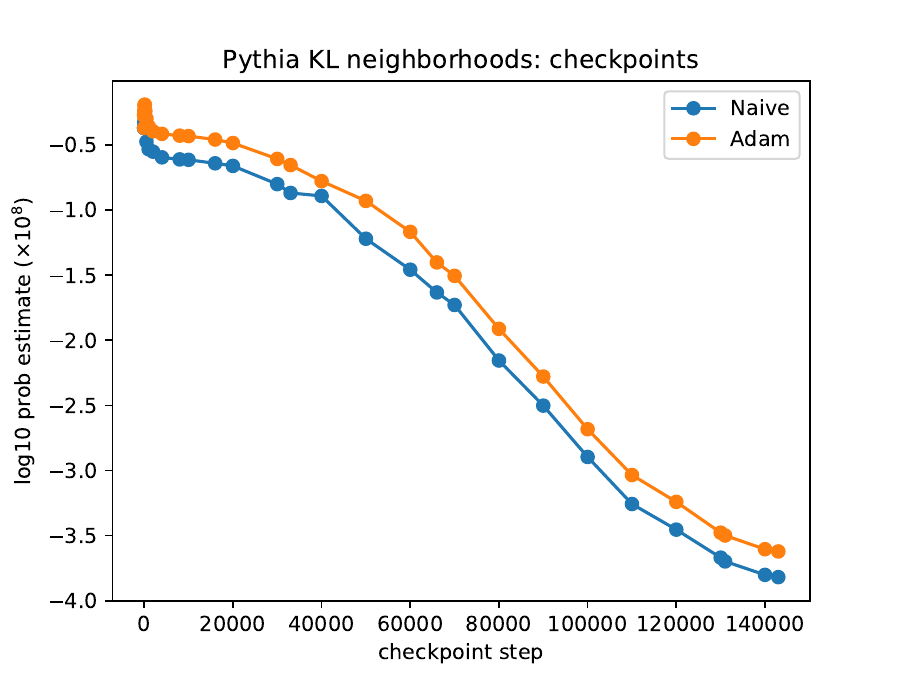}\\
    \includegraphics[width=\columnwidth]{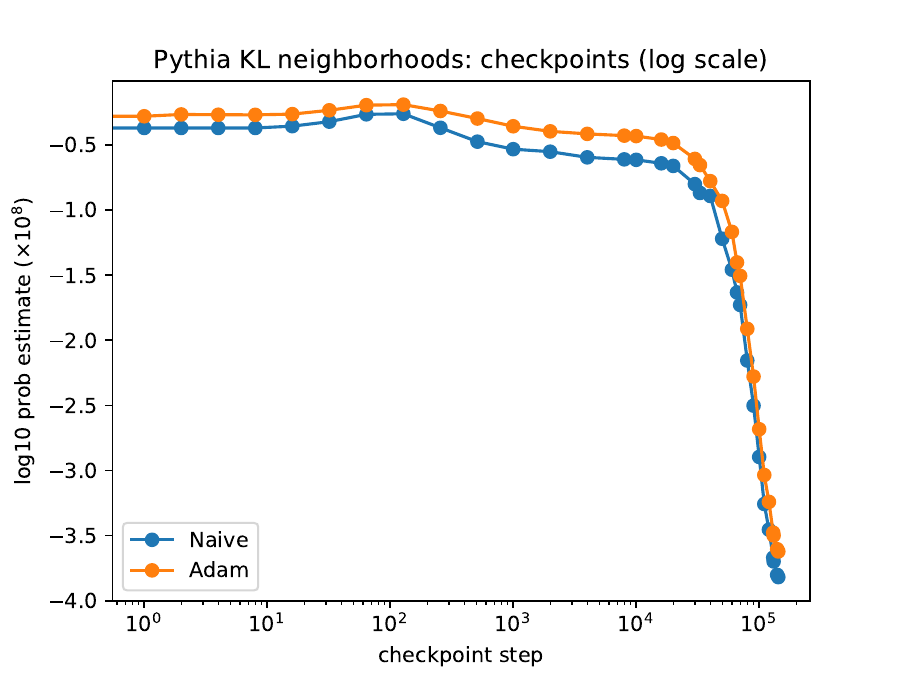}
    \caption{Local volume decrease while training Pythia 31M}
    \label{fig:pythia_chkpts}
\end{figure}
\begin{figure}[!ht]
    \centering
    \includegraphics[width=\columnwidth]{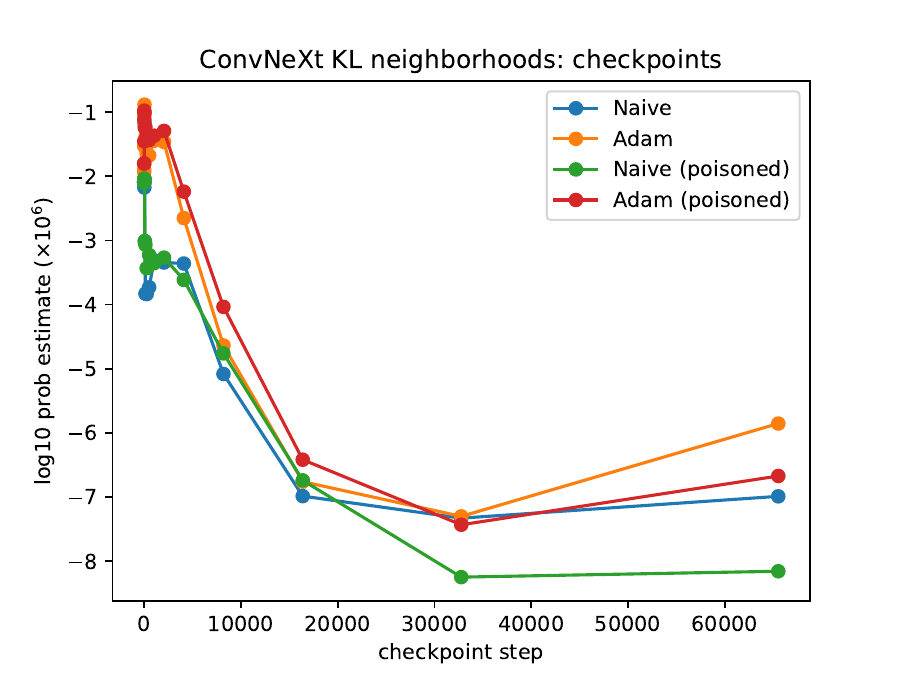}\\
    \includegraphics[width=\columnwidth]{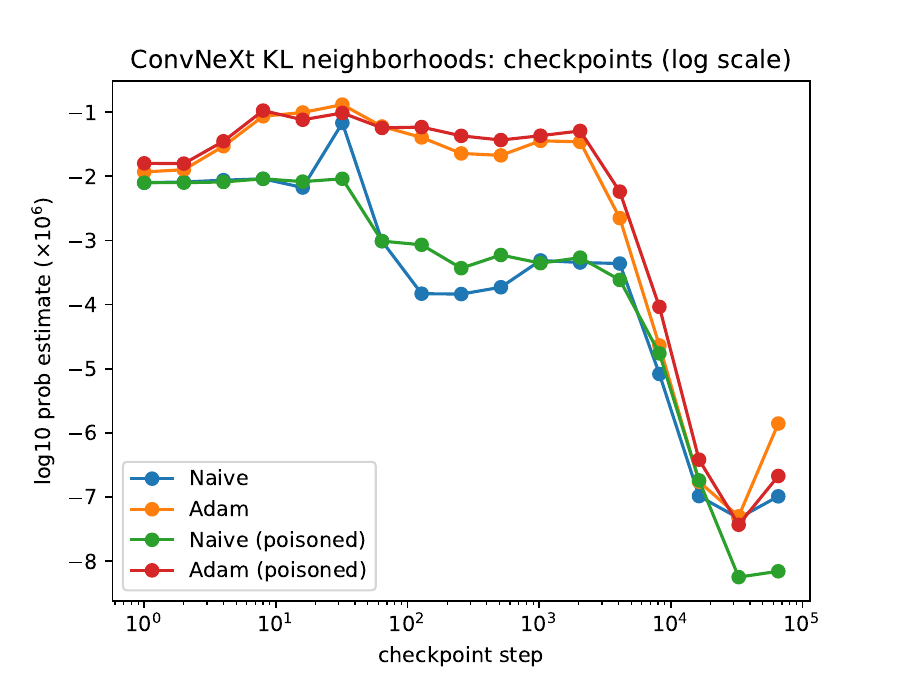}\\
    \hspace{-5mm}
    \includegraphics[width=.925\columnwidth]{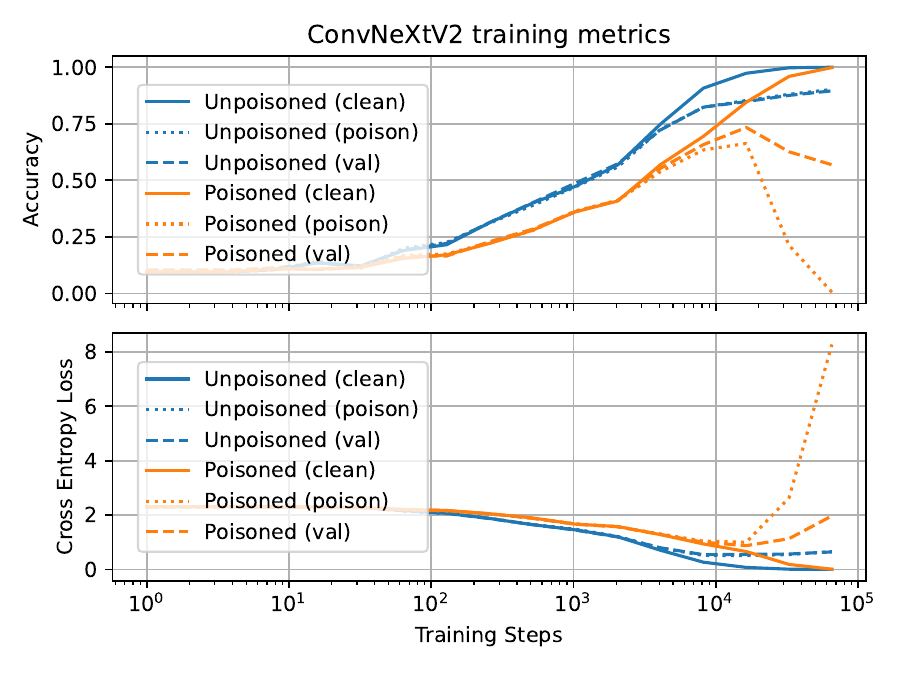}
    \caption{Local volume decrease while training ConvNeXt V2 Atto, and training metrics across datasets}
    \label{fig:convnext_chkpts}
\end{figure}

As expected, local volume tends to decrease during training, as the network learns more and its description length increases. For Pythia, this decrease is smooth and approximately exponential after an rapid drop early in training (Figure~\ref{fig:pythia_chkpts}). In this case, the Adam preconditioner yields modestly larger local volume estimates than the unpreconditioned method throughout training, but the two estimates follow parallel trends.

For ConvNeXt, the poisoned network actually has larger local volume for much of training, and then drops below the unpoisoned network around 30,000 steps, which is also when the val-set and poison-set losses diverge strongly from each other. This makes sense: early in  training, the poisoned loss is just holding back the network (worse loss across all three datasets), slowing the decrease in local volume. Later in training, the network overfits, decreasing its local volume to below the unpoisoned network's. This corresponds to a larger description length for the poisoned (overfit, poorly-generalizing) network.

\subsection{Across cutoffs}
\begin{figure}[!ht]
    \centering
    \includegraphics[width=\columnwidth]{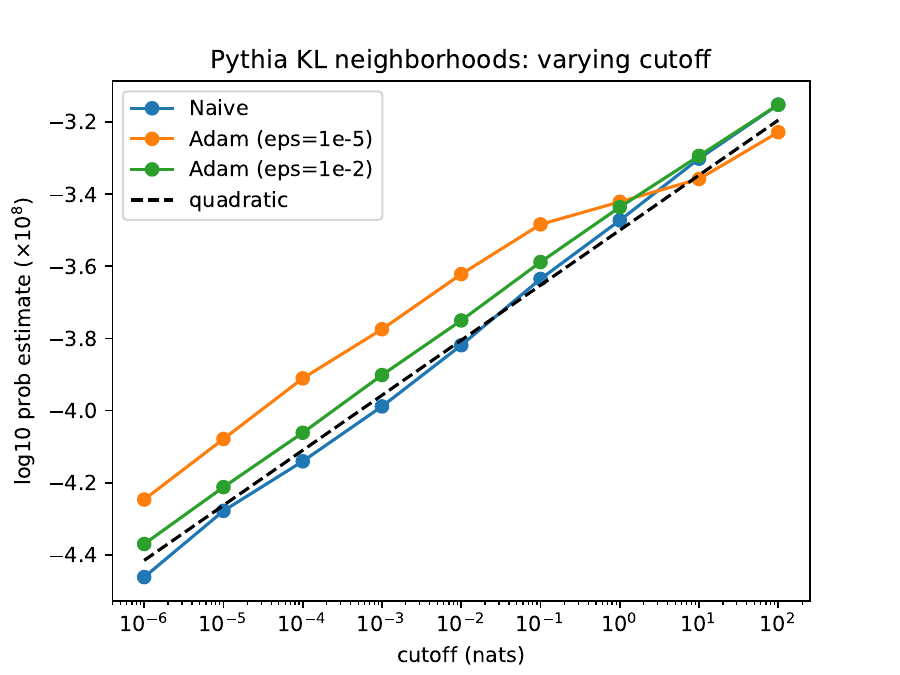}
    \caption{Results for various cutoffs on Pythia 31M}
    \label{fig:pythia_cutoff}
\end{figure}
\begin{figure}[!ht]
    \centering
    \includegraphics[width=\columnwidth]{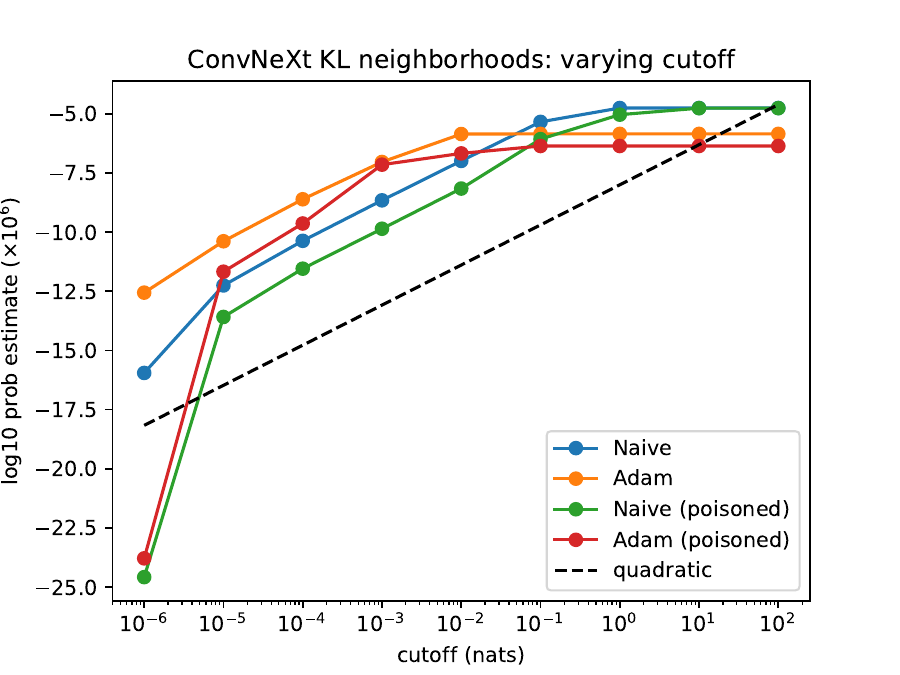}
    \caption{Results for various cutoffs on ConvNeXt V2 Atto}
    \label{fig:convnext_cutoff}
\end{figure}

When varying the cutoff, we see a roughly power-law trend for local volume on Pythia (\ref{fig:pythia_cutoff}). The log-log slope is consistent with $n/2$, where $n$ is the model dimension; this is what would be expected for a purely quadratic cost function. A line with this slope is shown in black for comparison. At very high cutoffs ($\geq10$ nats), the Adam preconditioner begins to fail, producing smaller-than-naive estimates, although raising our $\eps$ hyperparameter counteracts this. (Note that 10 nats corresponds to a perplexity of over 20,000, so this is a fairly extreme cutoff.)

For ConvNeXt, the result is similar for cutoffs between $10^{-5}$ and $10^{-2}$. The preconditioner again fails at high cutoff, sooner than for Pythia. At very low cutoffs ($10^{-6}$ nats), the poisoned network's local volume suddenly plummets. We have confirmed that this is a floating point precision issue: the radius-finding binary search fails to converge for cutoffs this low.

The consistent slope of $n/2$ for small cutoffs is in contrast with known results on the Local Learning Coefficient \citep{hoogland2024developmental, wang2024differentiation}, where this is a strict upper bound that is not seen in practice. We believe that this is due to our choice of KL over the more-singular training loss as a cost function, or our imposition of a Gaussian prior, which (by design) smoothly cuts off the measure of flat directions.\footnote{This could be tested by replacing the explicit Gaussian-integral computation with an additional quadratic term in the cost function.}


\section{Conclusion}

In this work, we introduced an efficient algorithm for estimating the probability that a network from some behaviorally-defined region would be sampled from a Gaussian or uniform prior, or equivalently, the network's \emph{local volume}. While the method is demonstrably more accurate than prior state of the art, it is still unclear how close our estimates are to the ground truth. Nevertheless, we find that our estimated local volume decreases with training time, and is smaller for networks that overfit than for generalizing networks, suggesting that it at least correlates with the true local volume.

Our results are broadly consistent with the volume hypothesis. As expected, badly-generalizing poisoned networks were observed to have smaller local volumes than well-generalizing unpoisoned ones. That said, more research is needed to confirm or refute any specific version of the volume hypothesis.

One promising direction for future work may be to use stochastic gradient Langevin dynamics (SGLD) to improve the importance-sampling step, by proposing directions along which to measure the neighborhood's radius.

Another possible application of SGLD is to replace our radius-finding estimator entirely, as is done in Singular Learning Theory research with the LLC estimator. SGLD does not directly give volume information, but thermodynamic integration can be used to compute a volume from a sequence of SGLD equilibria. This would potentially give a more accurate (albeit much slower) local-volume estimator, which could be used as a ground truth to evaluate the accuracy of our estimator.

We are excited to see practical applications of local volume estimation. We think it may be useful for predicting generalization performance. More speculatively, if we define the cost function to be the model's behavior on a relatively narrow distribution-- say, a set of math problems fed to a large language model-- the local volume may tell us something about how ``difficult'' these problems are for the model, or how hard it is ``thinking.'' We hope that the estimators we describe here can eventually be useful for detecting backdoors, scheming, or other unwanted hidden behavior in models.

Another possible direction may be to estimate the probability measure of neighborhoods around initializations that lead to a given final behavior after training, which corresponds almost exactly to the probability of SGD producing that trained behavior. This would allow for a precise quantitative evaluation of the volume hypothesis, and could potentially be accomplished via the training Jacobian \citep{belrose2024jacobian}.

\section*{Contributions and Acknowledgements}

Adam Scherlis came up with the main ideas of this paper, wrote the code and performed all experiments and data analysis, and did much of the writing. Nora Belrose proposed the idea of using expected KL divergence as a cost function in lieu of training loss, pointed out the connection with minimum description length, proposed a last-minute reframing of the paper in terms of ``sampling trained networks at random,'' and wrote various parts of the paper. We thank Jesse Hoogland, Dmitry Vaintrob, and others for useful conversations.

Adam and Nora are funded by a \href{https://www.openphilanthropy.org/grants/eleuther-ai-interpretability-research/}{grant} from Open Philanthropy. We thank Coreweave for computing resources.

\FloatBarrier

\section*{Impact Statement}

This paper presents work whose goal is to advance the science of deep learning. At this early stage of research the social impacts are uncertain and indirect, although we hope that future research building on this work may be used to enhance generalization and reliability of neural networks.

\bibliography{icml2025}
\bibliographystyle{icml2025}

\newpage
\appendix
\onecolumn

\section{Ellipsoidal toy model}
\label{app:toy_model}
In this appendix we collect theoretical results and analysis for the simple case of a quadratic cost function taking its minimum at the anchor point, or (slightly more generally) for an ellipsoidal neighborhood. This case is theoretically tractable and gives qualitatively accurate predictions for actual neighborhoods.

\subsection{Variance of the log-estimator}
\label{app:variance}

The variance of our local volume estimator is large when $S$ contains outlier directions which have a large effect on the volume. Intuitively, it is difficult to estimate the volume of a needle or pancake by measuring its size along uniformly sampled directions. Most samples will be far closer to the minimum than to the maximum radius. We can formalize this intuition in the following way. 

Consider an ellipsoid $S = \{ x \in 
\R^N : x^T A x \leq 1 \}$ for some p.s.d. matrix $A$. Assume also that our anchor $s_0$ is equal to the centroid of $S$. Now the radial function has the closed form:
\begin{equation} r(\vu)^{-2} = \vu^T \mathbf{A} \vu \end{equation}
It turns out that the variance of this quadratic form, assuming $\vu$ is uniformly distributed on the unit sphere, is
\begin{equation}
    \mathrm{Var}(\vu^T \mathbf{A} \vu) = \frac{2}{n + 2} \mathrm{Var}(\lambda),
\end{equation}
where $\mathrm{Var}(\lambda)$ is the variance of the eigenvalues of $\mathbf{A}$. Using a Taylor expansion around the mean, the variance of the log radial function is roughly half the squared \href{https://en.wikipedia.org/wiki/Coefficient_of_variation}{coefficient of variation} of the spectrum. As a result, the variance of the log-estimator is approximately:
\begin{equation}
    \mathrm{Var} \Big [ -\frac{n}{2} \log( \vu^T \mathbf{A} \vu ) \Big ] \approx \frac{n^3}{2(n + 2)} \cdot \frac{\mathrm{Var(\lambda)}}{\E[\lambda]^2}.
\end{equation}
If the spectrum of $\mathbf{A}$ has variance of order $\mathbb{E}[\lambda]^2$, as is generally the case for neural network Hessians, this variance will be extremely large (of order $n^2$ where $n$ is the parameter count).


\subsection{Underestimation and the Hessian spectrum}
\label{app:toy_hessian}

Consider an ellipsoidal basin, centered at the origin. Suppose the inverse principal axis radii $1/R_i$ vary smoothly over several orders of magnitude according to some probability distribution $P(1/R)$. Define a diagonal matrix $H$ s.t.
\eqn{
H_{ii} = 1/{R_i^2}
}

Then define
\eqn{
f(x) &:= \frac12 \sum_{ij} H_{ij}x_ix_j
}
so that $f(x) \le \frac12 \iff x\in X$.

Note that the distance $R(\hat v)$ to the boundary for a unit vector $\hat v$ is then given by
\eqn{
f(R(\hat v)\hat v) &= \frac12\\
R(\hat v)^2 f(\hat v) &= \frac12\\
R(\hat v)^2 &= \frac1{\sum_{ij}H_{ij}\hat v_i\hat v_j}\\
\frac1{R(\hat v)^2}&= {\sum_i \frac{\hat v_i^2}{R_i^2}}
}

If the dimension is high enough, such that $1/R_i$ stays nearly constant across $k\gg1$ consecutive axes, then the $k$ axes closest to some value $1/R$ will contribute to the sum according to
\eqn{
\sum_{i=n}^{n+k-1} \frac {\hat v_i^2}{R_i^2} &\approx \frac1{R^2}\sum_{i=n}^{n+k-1}{\hat v_i^2}\\
&\approx \frac1{R^2}\frac kn
}
where in the second line we have made use of the fact that $N(0,1/n)$ approximates a uniform unit sphere.

Then, summing along the distribution $P(1/R)$, we get
\eqn{
\frac1{R(\hat v)^2} &\approx \int \frac1{R^2} d P}
Therefore, in this limit, $R(\hat v)^2$ will approximate the harmonic mean of $R_i^2$ with high probability. This gives the wrong volume; the correct volume is $\propto\prod_i R_i$, so it is a function of the geometric mean of $R_i$. (Recall that the harmonic mean is bounded above by the geometric mean.) The naive volume estimator will still be unbiased (as it always is), but most samples will be close to a modal value that is significantly smaller than the true volume. The difference between mean and mode will be produced entirely by a heavy right tail, which would not be evident from a typical sample of a few random vectors.\\

When applying this analysis to real neural networks, there is an extra complication: some principal axes are outliers, which breaks the law-of-large-numbers approximation used above. However, this actually makes the underestimation problem worse in practice, since the tail (driven by the outliers) will be even heavier.\\


\subsection{Failure of strong volume hypothesis on small scales}
\label{app:sgd}
Consider a quadratic loss function with Hessian $\MH$. If the initialization distribution $\mu_0$ has covariance matrix $\mathbf{I}$, then at timestep $t$ the covariance is $\exp(-\MH t) \exp(-\MH t)^T$. Assuming $\mu_0$ is a zero-mean Gaussian, the log density of parameters $\theta$ at time $t$ is proportional to $\theta \exp(2 \MH t) \theta^T$, which is in general not proportional to the loss $\frac{1}{2} \theta \MH \theta^T$. The probability mass becomes concentrated along directions of higher curvature (larger Hessian eigenvalues) exponentially faster than along directions of lower curvature.

If we introduce isotropic noise and solve the resulting Fokker-Planck equation, it can be shown that the log-density instead converges to something proportional to the loss, as in SGLD. However, if the noise is not isotropic -- in particular if it is stronger in more steeply-curved directions, as is true in practice -- then this fails \citep{mandt2018sgd}.


\section{Another volume hypothesis}
\label{app:measure-vh}

Let $\mu_0$ be the probability measure on $\R^N$ from which the initial network parameters $\theta_0$ are sampled, usually a uniform distribution on a compact set or a Gaussian. Let $\mu_t$ be the distribution over network parameters at timestep $t$ in training, and let $f_t(x) = \frac{d\mu_t}{d\mu_0}$ be the probability density of parameters $x$ at time $t$.\footnote{Formally, the \href{https://en.wikipedia.org/wiki/Radon-Nikodym_theorem}{Radon-Nikodym derivative} of $\mu_t$ w.r.t. $\mu_0$. This quantity exists if $\mu_t$ is absolutely continuous w.r.t. $\mu_0$.} We can decompose the posterior probability of behaviorally distinct regions of parameter space, such as regions of low loss with differing degrees of generalization, as follows.

Let $A \subset \R^N$ and $B \subset \R^N$ be two disjoint regions of parameter space, both with consistently low training loss,\footnote{i.e. contained in a low-loss manifold \citep{benton2021loss}} but perhaps distinguished by their performance on a held-out test set. The probability that training will yield an element of $A$ can be decomposed as
\begin{align}
    \log \mathbb{P}(\theta \in A) &= \log \Big [ \mu_0(A) \cdot \frac{1}{\mu_0(A)} \int_{A} f_t d\mu_0 \Big ]\\
    &= \underbrace{\log \mu_0(A)}_\textbf{volume} + \underbrace{\log \mathbb{E}_{x \sim \mathrm{Unif}(A)}\big [ f_t(x) \big ]}_\textbf{mean density}
\end{align}
and the log probability ratio is
\begin{equation}\label{eq:volume-density}
    \log \frac{\mathbb{P}(\theta \in A)}{\mathbb{P}(\theta \in B)} = \underbrace{\log \frac{\mu_0(A)}{\mu_0(B)}}_\textbf{volume ratio} + \underbrace{\log \frac{\mathbb{E}_{x \sim \mu_0 \vert A}\big [ f_t(x) \big ]}{\mathbb{E}_{x \sim \mu_0 \vert B} \big [ f_t(x) \big ]}}_\textbf{density ratio},
\end{equation}
where $\mu_0 \vert A$ denotes the restriction of $\mu_0$ to $A$.\footnote{For example, if $\mu_0 = \mathrm{Unif}(S)$ for some compact $S \subset \R^N$, then $\mu_0 \vert A = \mathrm{Unif}(A \cap S)$. If $\mu_0$ is a Gaussian, then $\mu_0 \vert A$ is a truncated Gaussian with support $A$.} Note that at $t = 0$ we have $f_0(x) = \frac{d\mu_0}{d\mu_0}(x) = 1$ for any $x$, so that at early times $t$ the density ratio term in Eq.~\ref{eq:volume-density} should be small. 

The strong volume hypothesis would imply that the densities of $A$ and $B$ would be a function only of their training loss, so that the density ratio term would be zero at late times.

A restricted form of the volume hypothesis, for suitable choices of $A$ and $B$, is as follows: \textbf{Even at the end of training, the volume ratio term in Eq.\ref{eq:volume-density} should be larger than the density ratio term.}\footnote{
Of course, if the networks in $A$ and the networks in $B$ differ significantly in terms of their performance on the training set, the density ratio term must become very large as $t \rightarrow \infty$, since a well-tuned optimizer is guaranteed to bring the loss close to a local minimum. This is why we require $A$ and $B$ to be low-training-loss.}

\section{Naive estimator gives a lower bound on local volume}
\label{app:jensen_gap}
\paragraph{Underestimation problem.}

In practice, we estimate $\log \mathrm{vol}(S)$, rather than $\mathrm{vol}(S)$ itself, to prevent numerical overflow or underflow. \href{https://en.wikipedia.org/wiki/Jensen%27s_inequality}{Jensen's inequality} tells us that the logarithm of an unbiased estimator is a \emph{downwardly biased} estimator for the logarithm of the population parameter:
\begin{equation}
    \log \mathrm{vol}(S) \geq \E[\log \widehat{\mathrm{vol}(S)}],
\end{equation}
with equality if and only if the log-estimator is constant. 

This gap is especially large when the variance of the log-estimator is much larger than 1. For example, if the log-estimator is normally distributed with standard deviation $\sigma$,\footnote{As shown theoretically in Appendix~\ref{app:variance} and confirmed in our empirical results, the standard deviation in our case is in fact of order $n$, so this gap can be expected to be significant.} the gap is $\sigma^2/2$. In these cases, taking more samples will not necessarily fix the problem, because the vast majority of samples will underestimate by a similar amount. A more involved toy model of this phenomenon is given in Appendix~\ref{app:toy_hessian}.

Another way to phrase this same result is that the estimator for the volume itself (rather than its log) has a very heavy right tail.

\paragraph{Smooth maximum.} In practice, $n$ will be extremely large, ranging from $10^6$ to $10^{12}$ parameters. It is therefore worth considering the behavior of our estimator as in the large-$n$ limit. First note that
\begin{equation}
    \E [\log \widehat{\mathrm{vol}(S)}] \propto \E [\log \sum_{i = 1}^k \exp \big ( n \log r(\vu_i) \big )].
\end{equation}
LogSumExp is sometimes used as a continuous relaxation of the $\mathrm{max}$ function, because for any fixed set of values $\{ x_1, \ldots, x_k \}$ we have:
\begin{equation}\label{eq:realsoftmax}
    \lim_{n \rightarrow \infty} \frac{1}{n} \log \sum_{i = 1}^k \exp \big ( n x_i \big ) = \max(\{ x_1, \ldots, x_k \}).
\end{equation}
This suggests that, in the large-$n$ limit, the normalized log volume estimate $\frac{1}{n} \E [\log \widehat{\mathrm{vol}(S)}]$ will be close to the \emph{maximum} of our log-radius samples. Empirically, we find that this is already very nearly true for tiny networks of a few thousand parameters (Figure~\ref{fig:mlp_basins}).

\paragraph{Markov's inequality.} Since our estimator is a non-negative random variable, we can use \href{https://en.wikipedia.org/wiki/Markov%27s_inequality}{Markov's inequality} to show that with high probability, our estimate of the log-volume will not significantly \emph{overestimate} the true value:
\begin{equation}
    \mathbb{P}\Big ( \log \widehat{\mathrm{vol}(S)} - \log \mathrm{vol}(S) \geq \log k \Big ) \leq \frac{1}{k}
\end{equation}
That is, the probability that we overestimate the true volume by $m > 0$ orders of magnitude is at most one in $10^m$. As we expect theoretically and confirm empirically, our Monte Carlo samples for the volume vary over thousands or millions of orders of magnitude. Therefore, overestimating by $m=100$ orders of magnitude would be an imperceptible overestimate, and is extraordinarily unlikely (probability below $10^{-100}$). For this reason, we can treat our naive estimator as a very confident lower bound on the true local volume.

\section{Results on ConvNeXt training data}
\label{app:clean}

We report here results on ConvNeXt for local KL volume computed over the clean split of the training data (rather than the validation data as in Section~\ref{sec:results}).

\begin{figure}[!ht]
    \centering
    \includegraphics[width=\columnwidth]{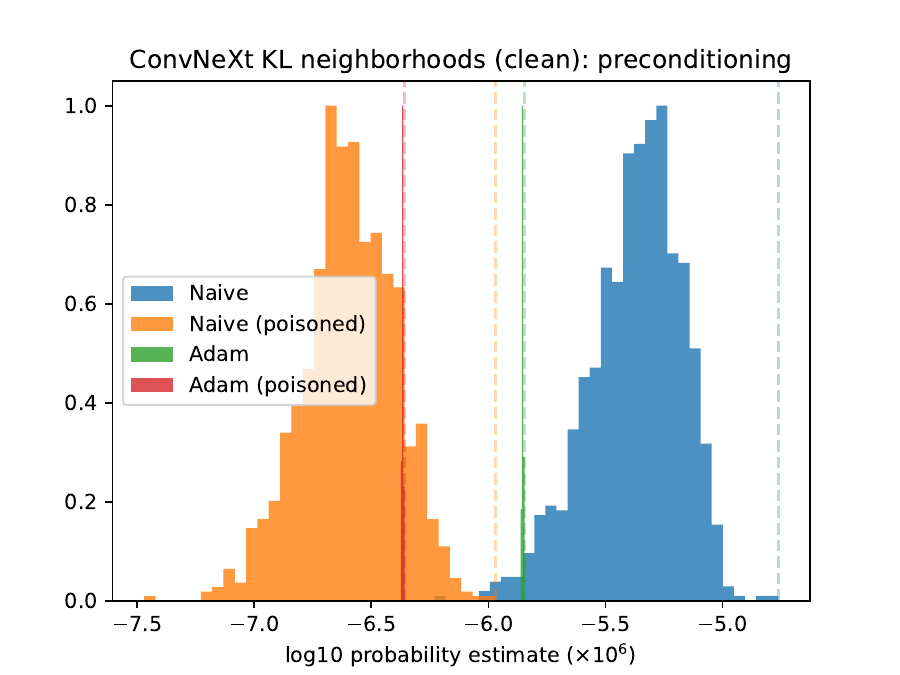}
    \caption{Results ($k=1000$) with and without Adam preconditioner on ConvNeXt Atto poisoned and unpoisoned}
    \label{fig:convnext_histo}
\end{figure}
\begin{figure}[!ht]
    \centering
    \includegraphics[width=\columnwidth]{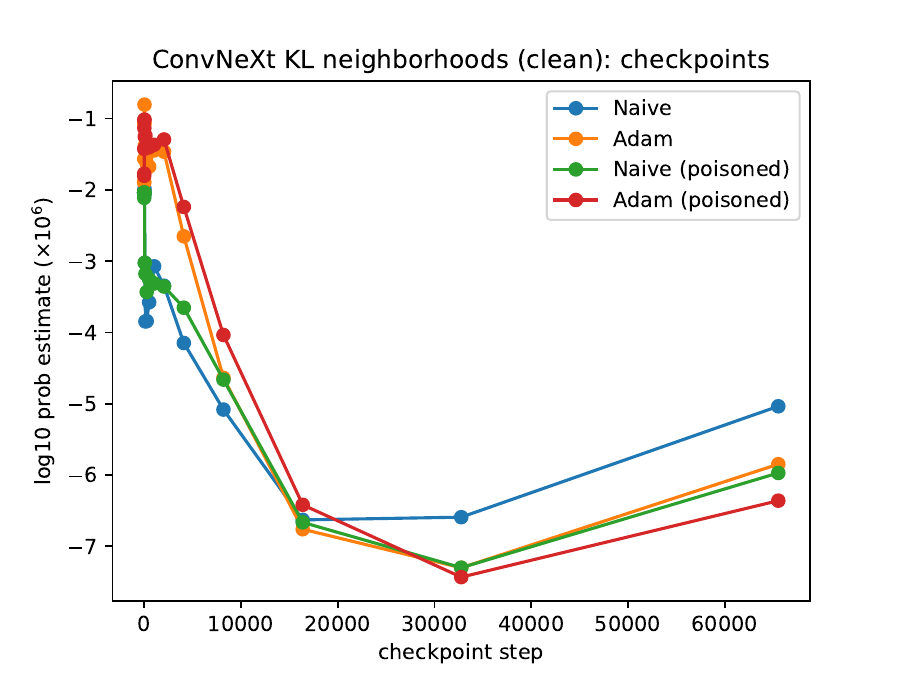}\\
    \includegraphics[width=\columnwidth]{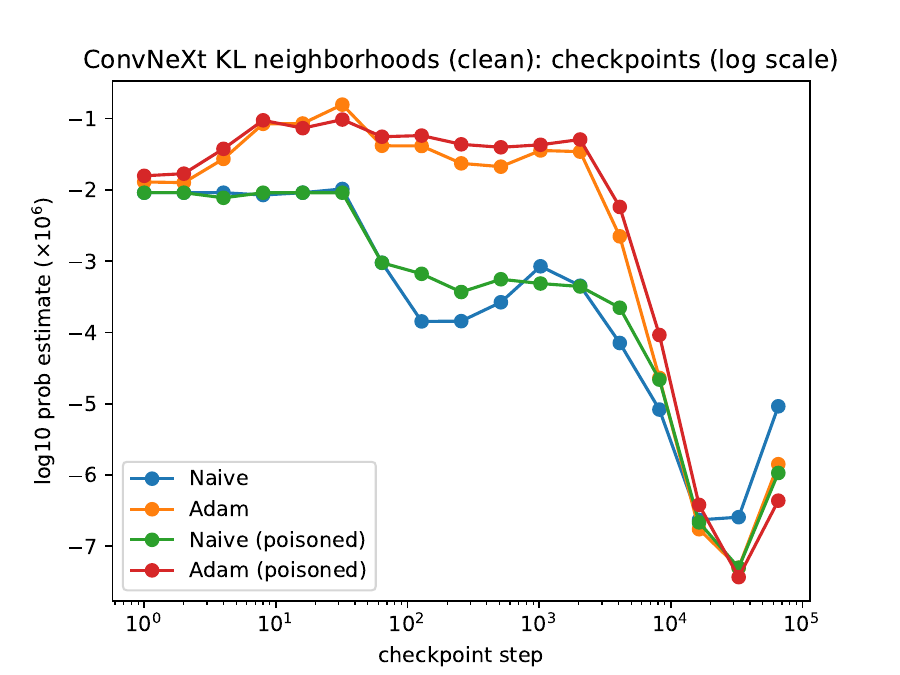}\\
    \caption{Local volume decrease while training ConvNeXt V2 Atto}
    \label{fig:convnext_chkpts}
\end{figure}

\end{document}